\documentclass[sigconf]{acmart}

\usepackage[linesnumbered,ruled]{algorithm2e}

\SetAlFnt{\small}
\SetAlCapFnt{\small}
\SetAlCapNameFnt{\small}
\IncMargin{-\parindent}
\usepackage{multirow}
\usepackage{graphicx}
\usepackage{subcaption}
\usepackage[export]{adjustbox}
\usepackage{balance}

\AtBeginDocument{%
  }

\copyrightyear{2026}
\acmYear{2026}
\setcopyright{cc}
\setcctype{by}
\acmConference[CIKM '26] {Proceedings of the 35th ACM International Conference on Information and Knowledge Management}{November 7--11, 2026}{Rome, Italy.}
\acmBooktitle{Proceedings of the 35th ACM International Conference on Information and Knowledge Management (CIKM '26), November 7--11, 2026, Rome, Italy}
\acmISBN{979-8-4007-2539-5/2026/11}
\acmDOI{10.1145/3799682.3840860}

\begin{document}

\title{HiTS-CL: A Continual Learning Framework for Long-Horizon Temporal Knowledge Graph Extrapolation}

\author{Yansong Liu}
\email{liuyansong@buaa.edu.cn}
\orcid{0000-0002-6260-5289}
\affiliation{%
  \institution{Beihang University}
  \city{Beijing}
  \country{China}
}

\author{Rui Liu}
\email{lr@buaa.edu.cn}
\orcid{0000-0003-1373-6108}
\affiliation{%
  \institution{Beihang University}
  \city{Beijing}
  \country{China}
}

\author{Yuan Zuo}
\authornote{corresponding authors.}
\email{zuoyuan@buaa.edu.cn}
\orcid{0000-0001-8516-2567}
\affiliation{%
  \institution{Beihang University}
  \city{Beijing}
  \country{China}
}

\author{Hongwei Zhao}
\email{zhaohongwei@buaa.edu.cn}
\orcid{0000-0001-6821-4185}
\affiliation{%
  \institution{Beihang University}
  \city{Beijing}
  \country{China}
}

\author{Da Fu}
\email{fuda@buaa.edu.cn}
\orcid{0009-0004-3991-0623}
\affiliation{%
  \institution{Beihang University}
  \city{Beijing}
  \country{China}
}

\author{Fuwei Zhang}
\email{fuwei.zhang@foxmail.com}
\orcid{0000-0002-7711-866X}
\affiliation{%
  \institution{Beihang University}
  \city{Beijing}
  \country{China}
}

\author{Fuzhen Zhuang}
\email{zhuangfuzhen@buaa.edu.cn}
\orcid{0000-0001-9170-7009}
\affiliation{%
  \department{School of Artificial Intelligence}
  \institution{Beihang University}
  \city{Beijing}
  \country{China}
}

\author{Yong Chen}
\email{alphawolf.chen@gmail.com}
\orcid{0000-0002-1348-9218}
\affiliation{%
  \institution{Beijing University of Posts and Telecommunications}
  \city{Beijing}
  \country{China}
}

\author{Zhe Li}
\email{lizhe\_hbeu@vip.163.com}
\authornotemark[1]
\orcid{0000-0003-1680-1083}
\affiliation{%
  \institution{Hubei Engineering University}
  \city{Xiaogan}
  \state{Hubei}
  \country{China}
}

\renewcommand{\shortauthors}{Yansong Liu et al.}

\begin{abstract}
Extrapolative temporal knowledge graph reasoning (TKGR) predicts future facts from historical snapshots. Most existing methods train once on an early prefix of the timeline and then use a frozen model for all future timestamps. We argue that this fixed-prefix protocol is misaligned with extrapolation. It learns from a static prefix, whereas the target stream is non-stationary: new entities and facts emerge, temporal dependencies shift across regimes, and recurring historical signals must be refreshed online. As a result, models trained only on early snapshots become outdated and degrade over long horizons. We address this mismatch by formulating extrapolative TKGR as continual learning over streaming snapshots. Under this view, effective extrapolation must jointly handle \emph{current dynamics}, \emph{stable knowledge}, and \emph{recurring historical evidence}. Based on these requirements, we propose \textbf{Hi}story-enhanced \textbf{T}wo-\textbf{S}tep \textbf{C}ontinual \textbf{L}earning (\textbf{HiTS-CL}), a backbone-agnostic continual learning framework for extrapolative TKGR. HiTS-CL tracks current dynamics via continual fine-tuning, preserves stable knowledge via multi-teacher adaptive distillation, and retains recurring historical evidence via a selective memory of recent and frequent facts. We integrate HiTS-CL into five representative TKGR backbones and evaluate it on four benchmark datasets. HiTS-CL consistently improves extrapolation accuracy, reduces long-horizon degradation, and outperforms strong continual-learning baselines, including a recent method for temporal knowledge graphs. Source code and data are available at \url{https://github.com/liuyansong98/HiTS-CL}.
\end{abstract}



\maketitle

\section{Introduction}
\label{sec:introduction}

Temporal Knowledge Graph Reasoning (TKGR) aims to infer missing temporal facts from evolving knowledge graphs represented as quadruples $(s,r,o,t)$ \cite{TKG_suvery2,TKG_suvery3, zhang2022along, zhang2026multi-granularity}. In \emph{extrapolative} TKGR, the goal is to predict facts at future timestamps beyond the observed timeline \cite{RE-Net}. This setting is important for applications such as temporal question answering \cite{QA_for_TKG1, QA_for_TKG2, QA_for_TKG3, QA_for_TKG4}, recommendation \cite{TKG_rec1, TKG_rec2}, information retrieval \cite{TKG_IR1}, and decision support \cite{TKG_medical_diagnosis, chaturvedi2024temporal}.

\begin{figure}[!t]
\centering
\subfloat[Performance Decay]{\includegraphics[width=0.45\columnwidth]{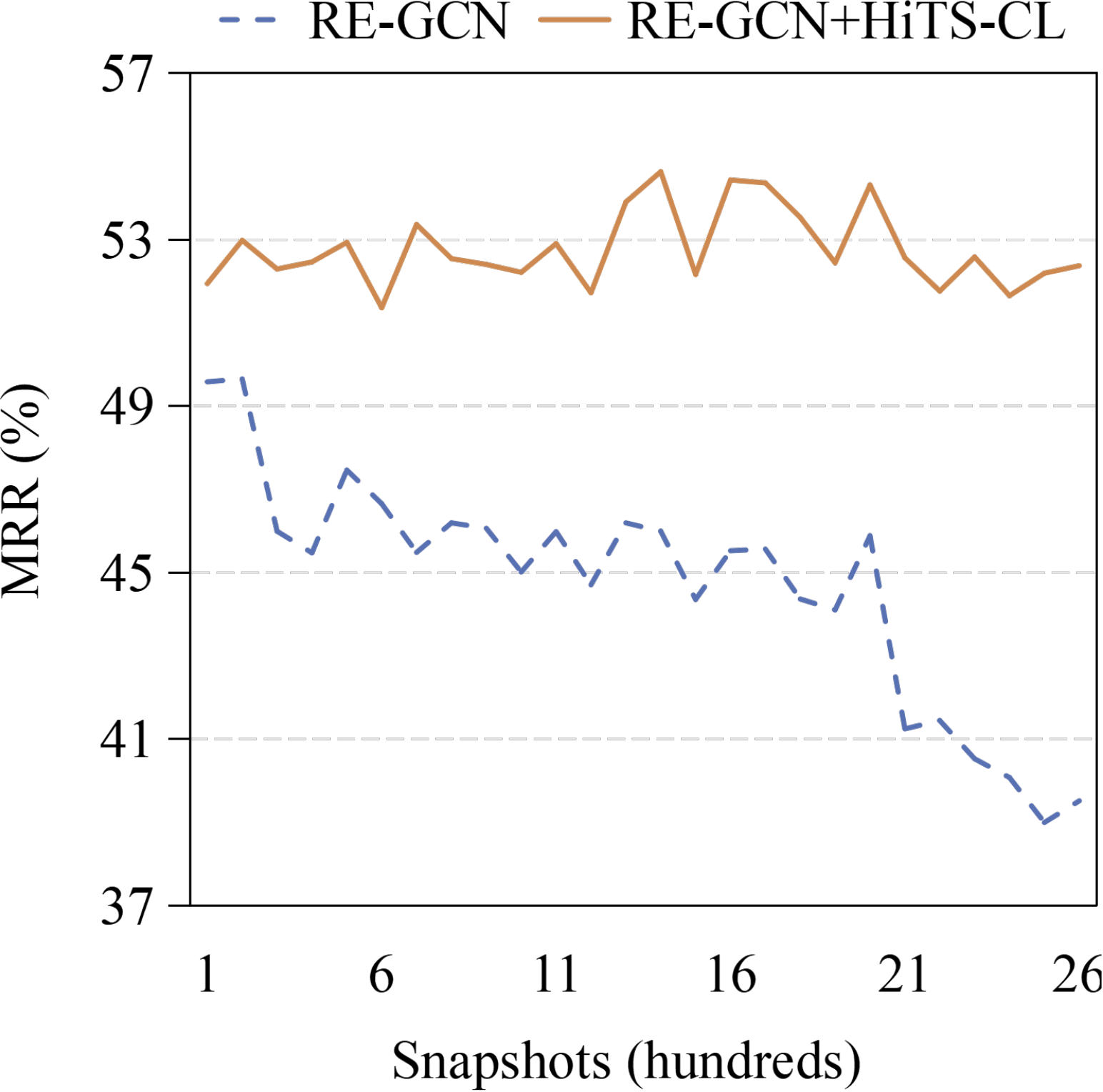}\label{fig:result_decay_RE-GCN}}
\hfill
\subfloat[Training Comparison]{\includegraphics[width=0.45\columnwidth]{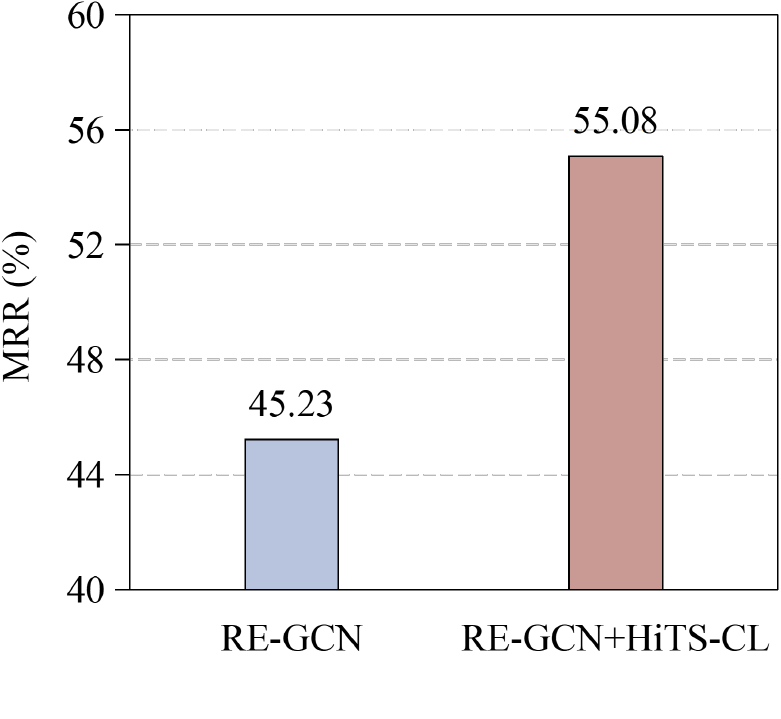}\label{fig:training_comparison_inro}}
\caption{(a) Long-horizon performance decay of a fixed-prefix backbone versus HiTS-CL on the test stream (MRR averaged over 100 snapshots). Dashed/solid curves denote the backbone and its HiTS-CL variant. (b) Fixed-prefix training versus continual updating evaluated on the immediately following test snapshot after the training period (snapshot 1001; train: 1--1000).}
\label{fig:result_decay}
\end{figure}

Most existing extrapolative TKGR methods follow a \emph{fixed-prefix} protocol: they are trained once on an early prefix of the timeline and then applied unchanged to all future timestamps, as in RE-GCN \cite{RE-GCN}, TANGO \cite{TANGO}, CEN \cite{CEN}, TiTer \cite{TITer}, HGLS \cite{HGLS}, RETIA \cite{RETIA}, LogCL \cite{LogCL}, TiPNN \cite{TiPNN}, and DiMNet \cite{DiMNet}. We argue that this protocol is fundamentally misaligned with extrapolation. It learns a static predictor from an early prefix, while the target stream is non-stationary and continues to evolve after deployment. As a result, models trained only on early snapshots exhibit clear long-horizon degradation (Figure~\ref{fig:result_decay_RE-GCN}) and can even underperform continually updated models on the first test snapshot immediately after training (Figure~\ref{fig:training_comparison_inro}).

This mismatch stems from three challenges. First, temporal knowledge graphs evolve in an open-world manner: new facts continually emerge. Second, temporal dependencies may shift across regimes: fixed-prefix models tend to capture stable knowledge but fail to adapt to transient current dynamics, as illustrated by the immediate performance drop in Figure~\ref{fig:training_comparison_inro}. Third, future prediction often depends on recurring historical evidence, such as periodic or high-frequency interaction patterns \cite{CyGNet, LogCL, HisRepeats, TiRGN}, which must be refreshed online without storing the entire history.

These observations motivate a different formulation: extrapolative TKGR should be treated as continual learning over streaming snapshots. Under this view, effective extrapolation must jointly capture \emph{current dynamics}, \emph{stable knowledge}, and \emph{recurring historical evidence}. This setting also differs from prior continual TKG formulations, which typically focus on within-snapshot completion rather than strict next-step forecasting under a no-leakage extrapolation protocol.

Based on this view, we propose \textbf{Hi}story-enhanced \textbf{T}wo-\textbf{S}tep \textbf{C}ontinual \textbf{L}earning (\textbf{HiTS-CL}), a \textbf{model-agnostic continual learning framework} for extrapolative TKGR. HiTS-CL tracks current dynamics through continual fine-tuning, preserves stable knowledge through multi-teacher adaptive distillation, and retains recurring historical evidence through a selective memory inspired by cache replacement strategies \cite{ARC}. In this way, HiTS-CL explicitly separates and recombines the three predictive ingredients required for extrapolation.

We integrate HiTS-CL into five representative TKGR backbones and evaluate it on four benchmark datasets. HiTS-CL consistently improves extrapolation accuracy, reduces long-horizon degradation, and outperforms strong continual-learning baselines as well as a recent continual-learning method for temporal knowledge graphs adapted to our no-leakage extrapolation setting. Our contributions are summarized as follows:
\begin{itemize}
    \item We identify a fundamental limitation of the fixed-prefix protocol for extrapolative TKGR: it relies on a static predictor for a non-stationary forecasting problem, leading to outdated models and long-horizon degradation.
    \item We reformulate extrapolative TKGR as continual learning over streaming snapshots, yielding a setting better aligned with evolving temporal knowledge graph streams.
    \item We propose \textbf{HiTS-CL}, a model-agnostic continual learning framework for extrapolative TKGR, which explicitly separates and recombines current dynamics, stable knowledge, and recurring historical evidence through continual fine-tuning, multi-teacher adaptive distillation, and selective-memory history enhancement.
    \item Experiments across multiple backbones and benchmark datasets show that HiTS-CL consistently improves extrapolation accuracy and outperforms strong baselines.
\end{itemize}

\section{Related Work}

\subsection{Temporal Knowledge Graph Reasoning}

Temporal Knowledge Graph Reasoning (TKGR) is commonly studied under interpolation and extrapolation settings \cite{RE-Net}. Given training timestamps $[0,T]$, interpolation predicts missing facts within the observed range ($0<t<T$), whereas extrapolation predicts facts at future timestamps ($t>T$). Extrapolation is generally more challenging and more relevant in practice. Existing methods model temporal evolution through structural and sequential patterns, such as snapshot-based graph encoders and temporal sequence modeling \cite{RE-GCN,CEN,RETIA,LogCL,DiMNet}. In parallel, recurring and periodic events have also been shown to provide strong signals for temporal forecasting \cite{CyGNet, LogCL, CENET, HIP, HisRepeats, RPC, TiRGN, DyMemR, zhang2024multi-view}. 

Despite these advances, most extrapolative TKGR methods still follow a fixed-prefix training protocol: they are trained once on early snapshots and then applied unchanged to future timestamps. As a result, they are not designed to adapt to evolving temporal streams during deployment. Our work differs in treating extrapolative TKGR as continual learning over streaming snapshots.

\begin{figure*}[t!]
    \centering 
    \includegraphics[width=\textwidth]{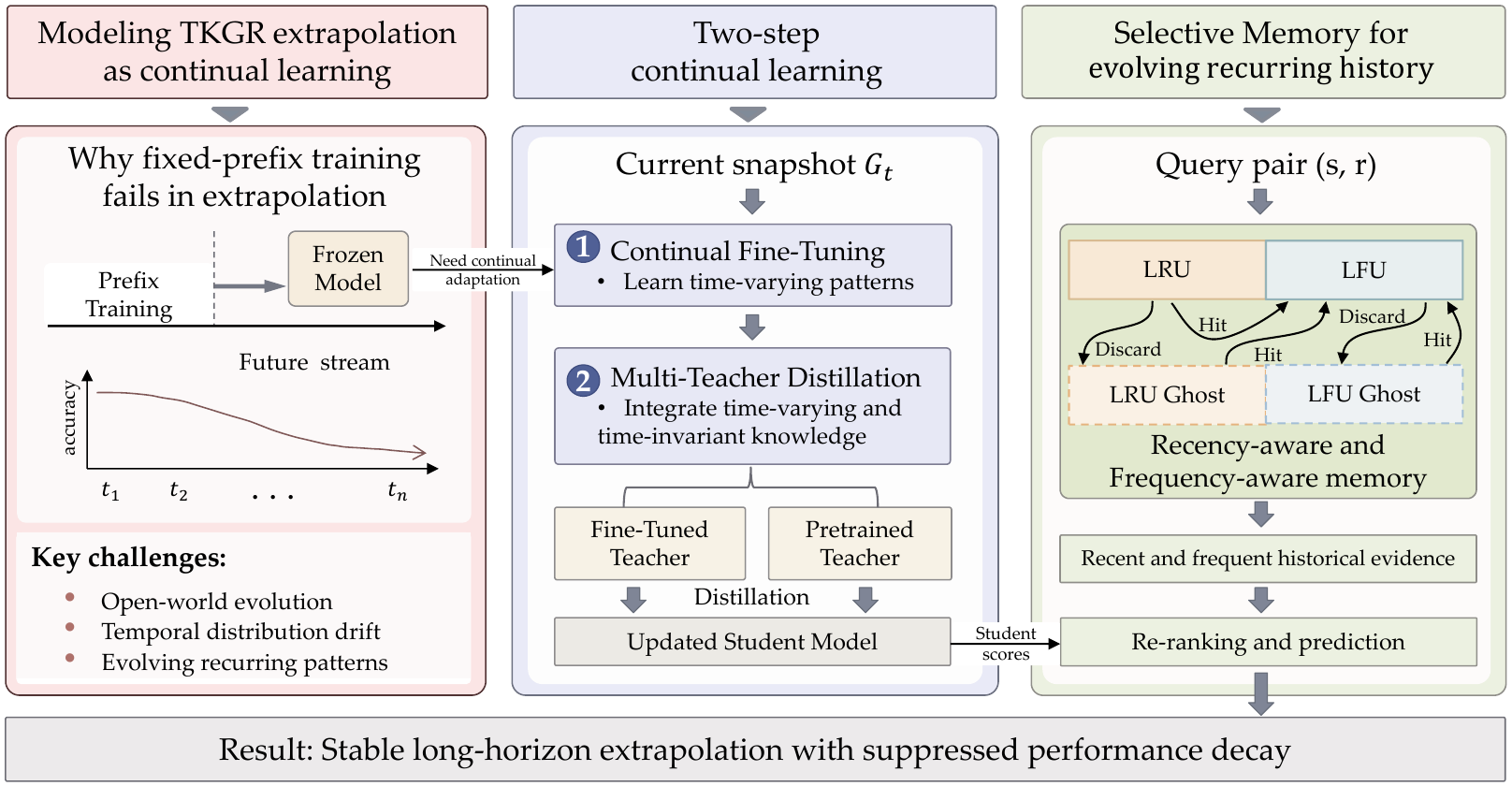}
    \caption{Overview of HiTS-CL: continual fine-tuning captures current dynamics, multi-teacher distillation preserves stable knowledge while incorporating time-varying patterns, and selective-memory history enhancement provides recurring historical evidence.}
    \label{fig:method_framework}
\end{figure*}

\subsection{Continual Learning and Continual Graph Learning}

Continual learning (CL) studies how models learn from sequential data while balancing adaptation and knowledge retention \cite{CL_survey}. It is also referred to as lifelong learning \cite{liu2017lifelong} or incremental learning \cite{castro2018end}. While CL has been extensively studied on Euclidean data \cite{CL_survey}, extending it to graphs introduces additional challenges due to evolving nodes, edges, and their propagated influence \cite{tian2024continual}. Representative graph continual learning approaches include IncDE \cite{IncDE}, SSRM \cite{SSRM}, and LKGE \cite{LKGE}.

Several works have extended continual learning to temporal knowledge graphs \cite{TKGC_cl_1, TKGC_cl_3, TKGC_cl_4}. However, they typically formulate learning within individual snapshots or incremental temporal KG completion settings, rather than next-step forecasting under an extrapolative protocol. In particular, their evaluation settings do not explicitly target streaming prediction where training and validation strictly precede future test snapshots. Among them, DGAR~\cite{TKGC_cl_4} is the closest to our work, as it also studies continual learning for temporal knowledge graphs. However, DGAR focuses on within-snapshot interpolation and delayed task-level evaluation: training, validation, and test facts are split within the same snapshot, and each task is evaluated only after incremental training over all tasks is completed. In contrast, HiTS-CL targets strict streaming extrapolation, where training and validation precede future test snapshots and evaluation is performed immediately after each continual update.

More broadly, most continual graph learning benchmarks focus on node or graph classification \cite{febrinanto2023graph}, whereas extrapolative TKGR is a temporal link prediction problem with many incremental stages. This makes continual extrapolation in temporal knowledge graphs a distinct and still under-explored setting.

\section{Preliminary and Problem Definition}

\subsection{Extrapolative Temporal Knowledge Graph Reasoning}

A temporal knowledge graph (TKG) is represented as a sequence of snapshots
$\{\mathcal{G}_t\}_{t=1}^T$.
Each snapshot $\mathcal{G}_t=(\mathcal{E}_t,\mathcal{R},\mathcal{Q}_t)$ is a directed multi-relational graph, where $\mathcal{E}_t$ and $\mathcal{R}$ denote the entity and relation sets, and $\mathcal{Q}_t=\{(s,r,o,t)\mid s,o\in\mathcal{E}_t,\ r\in\mathcal{R}\}$ is the set of facts observed at time $t$.

We consider TKGR under extrapolation, where the goal is to predict facts at future timestamps beyond the observed history. Given a query $(s,r,?,t)$, a TKGR model outputs scores over candidate entities in $\mathcal{E}_t$. We denote the score assigned to a candidate entity $o\in\mathcal{E}_t$ by $\hat{\psi}_{(s,r,?,t)}(o)$. Subject prediction is handled by introducing inverse relations, i.e., by predicting $(o,r^{-1},?,t)$.

\subsection{Extrapolative TKGR as Continual Learning}

Most existing extrapolative TKGR methods follow a fixed-prefix protocol: they are trained on an early prefix of the timeline and then kept unchanged for all future timestamps. This protocol is mismatched to non-stationary temporal streams. We therefore reformulate extrapolative TKGR as continual learning over streaming snapshots.

Starting from deployment time $t_0$, we define a sequence of continual learning stages
$\{\mathcal{T}_t\}_{t=t_0+1}^{T-1}$, where
\begin{equation}
\begin{aligned}
&\mathcal{T}_t = \left(\mathcal{D}_{\mathrm{tr}}^t,\mathcal{D}_{\mathrm{va}}^t,\mathcal{D}_{\mathrm{te}}^t\right),\\
\mathcal{D}_{\mathrm{tr}}^t = \mathcal{G}_t &,\;
\mathcal{D}_{\mathrm{va}}^t = \mathcal{G}_{t-1},\;
\mathcal{D}_{\mathrm{te}}^t = \mathcal{G}_{t+1}.
\end{aligned}
\end{equation}
At stage $t$, the model is updated on the current snapshot $\mathcal{G}_t$ and evaluated on the next snapshot $\mathcal{G}_{t+1}$, matching the streaming extrapolation setting without full retraining.

We use the previous snapshot $\mathcal{G}_{t-1}$ for validation for two reasons. First, unlike a within-snapshot hold-out strategy, it preserves the full current snapshot $\mathcal{G}_t$ for training, avoiding potential data insufficiency in the current update step. Second, we empirically find that this choice yields slightly better results across datasets (Section~\ref{sec:Validation_Strategy}).

\section{Method}

We propose \textbf{HiTS-CL}, a model-agnostic continual learning framework for extrapolative TKGR. Given a pretrained TKGR backbone, HiTS-CL incrementally adapts it to streaming snapshots in a stage-wise manner, as illustrated in Figure~\ref{fig:method_framework}. At each stage $t$, HiTS-CL first performs \textbf{continual fine-tuning} to capture \emph{current dynamics}, then applies \textbf{multi-teacher distillation} to preserve \emph{stable knowledge} while incorporating newly acquired time-varying patterns. In addition, it maintains a \textbf{selective memory} of historical facts, updated online by recency and frequency, to provide \emph{recurring historical evidence} for prediction.

\subsection{Two-Step Continual Learning}

Let $f_{\phi}(s,r,t)\in\mathbb{R}^{|\mathcal{E}_t|}$ denote a TKGR backbone parameterized by $\phi$, which outputs logits over candidate entities for a query $(s,r,?,t)$. The score of a candidate entity $o\in\mathcal{E}_t$ is denoted by $\hat{\psi}_{(s,r,?,t)}(o)$, i.e., the $o$-th entry of $f_{\phi}(s,r,t)$.

Before deployment, we pretrain the backbone on the offline training split and obtain a \textbf{pretrained teacher} $f^{\text{pre}}_{\theta}$. The parameters $\theta$ remain frozen during continual learning and serve as a stable reference for preserving \emph{stable knowledge}. During streaming, at each stage $t$, we maintain a \textbf{fine-tuned teacher} $f^{\text{ft}}_{\eta_t}$ and a \textbf{student} $f^{\text{stu}}_{\omega_t}$.

\subsubsection{Step 1: Continual Fine-Tuning}

At stage $t$, we update the fine-tuned teacher using only the newly arrived snapshot $\mathcal{G}_t$:
\begin{equation}
f^{\text{ft}}_{\eta_t}
=
\operatorname{FineTune}\!\left(f^{\text{ft}}_{\eta_{t-1}},\ \mathcal{G}_t\right).
\label{eq:fine-tuning}
\end{equation}
This step prioritizes plasticity, enabling rapid adaptation to current time-varying patterns. We use the preceding snapshot $\mathcal{G}_{t-1}$ for early stopping based on MRR, which follows the no-leakage extrapolation protocol and relies only on past data. Further analysis of this validation strategy is provided in Section~\ref{sec:Validation_Strategy}.

\subsubsection{Step 2: Multi-Teacher Adaptive Distillation}

Continual fine-tuning improves adaptation to time-varying patterns but may forget stable knowledge acquired during pretraining. We therefore distill from two teachers into a student: the pretrained teacher $f^{\text{pre}}_{\theta}$ provides stable knowledge, while the fine-tuned teacher $f^{\text{ft}}_{\eta_t}$ provides current time-varying knowledge. We introduce adaptivity in three components: student initialization, sample routing, and temperature scaling.

\paragraph{Student initialization}
We estimate teacher reliability on the current snapshot using MRR:
\begin{equation}
\begin{aligned}
m^{\text{pre}}_t &= \operatorname{MRR}\!\left(f^{\text{pre}}_{\theta},\ \mathcal{G}_{t}\right),\\
m^{\text{ft}}_t  &= \operatorname{MRR}\!\left(f^{\text{ft}}_{\eta_t},\ \mathcal{G}_{t}\right).
\end{aligned}
\end{equation}
We then compute a reliability weight
\begin{equation}
\rho_t=\frac{m^{\text{pre}}_t}{m^{\text{pre}}_t+m^{\text{ft}}_t},
\label{eq:rho}
\end{equation}
and initialize the student by reliability-weighted interpolation:
\begin{equation}
\omega_t^{-}=\rho_t\,\theta + (1-\rho_t)\,\eta_t.
\end{equation}
This initialization allows the student to start closer to the more reliable teacher on the current stage.

\paragraph{Sample routing}
Different teachers can be more reliable on different instances. We perform distillation on the current-stage training data $\mathcal{D}_{\mathrm{tr}}^t=\mathcal{G}_t$, which reflects the current regime. For each instance $(x,y)\in\mathcal{D}_{\mathrm{tr}}^t$, where $x$ is a query and $y$ is the ground-truth entity index, let
\begin{equation}
\psi^{\text{pre}}(x)=f^{\text{pre}}_{\theta}(x),\qquad
\psi^{\text{ft}}(x)=f^{\text{ft}}_{\eta_t}(x)
\end{equation}
denote the teacher logits. We define the per-sample supervised losses as
\begin{equation}
\ell^{\text{pre}}(x,y)=\operatorname{CE}\!\left(\psi^{\text{pre}}(x),y\right),\;
\ell^{\text{ft}}(x,y)=\operatorname{CE}\!\left(\psi^{\text{ft}}(x),y\right).
\end{equation}
We route each sample to the teacher with lower loss:
\begin{equation}
\begin{aligned}
\mathcal{D}^{\text{pre}}_t &= \{(x,y)\in\mathcal{D}_{\mathrm{tr}}^t \mid \ell^{\text{pre}}(x,y)\le \ell^{\text{ft}}(x,y)\},\\
\mathcal{D}^{\text{ft}}_t  &= \{(x,y)\in\mathcal{D}_{\mathrm{tr}}^t \mid \ell^{\text{ft}}(x,y)< \ell^{\text{pre}}(x,y)\}.
\end{aligned}
\end{equation}
This allows each teacher to supervise the student on the instances for which it is more reliable. 

\paragraph{Temperature scaling}
We use a per-sample temperature to reflect teacher confidence. For the routed teacher logits $\psi(x)$, let
\begin{equation}
p(x)=\operatorname{softmax}(\psi(x))
\end{equation}
and define the entropy
\begin{equation}
H(x)=-\sum_i p_i(x)\log\!\bigl(p_i(x)+\epsilon\bigr).
\end{equation}
We set
\begin{equation}
\tau_x=\alpha \cdot \operatorname{sigmoid}(H(x)),
\label{eq:temp}
\end{equation}
where $\epsilon$ is a small constant for numerical stability and $\alpha > 0$ modulates the differentiation among samples. Lower entropy yields smaller $\tau_x$ and thus sharper supervision.

\paragraph{Distillation loss}
Let
\begin{equation}
\begin{aligned}
p^{\text{pre}}_x &= \operatorname{softmax}\!\left(\psi^{\text{pre}}(x)/\tau_x\right),\\
p^{\text{ft}}_x  &= \operatorname{softmax}\!\left(\psi^{\text{ft}}(x)/\tau_x\right),\\
p^{\text{stu}}_x &= \operatorname{softmax}\!\left(\psi^{\text{stu}}(x)/\tau_x\right),
\end{aligned}
\end{equation}
where
\begin{equation}
\psi^{\text{stu}}(x)=f^{\text{stu}}_{\omega_t}(x).
\end{equation}
We distill via KL divergence on the routed subsets:
\begin{equation}
\begin{aligned}
\mathcal{L}^{\text{pre}}_{\text{distill}}
&=
\frac{1}{|\mathcal{D}^{\text{pre}}_t|}
\sum_{(x,y)\in\mathcal{D}^{\text{pre}}_t}
\mathrm{KL}\!\left(p^{\text{pre}}_x \,\|\, p^{\text{stu}}_x\right),\\
\mathcal{L}^{\text{ft}}_{\text{distill}}
&=
\frac{1}{|\mathcal{D}^{\text{ft}}_t|}
\sum_{(x,y)\in\mathcal{D}^{\text{ft}}_t}
\mathrm{KL}\!\left(p^{\text{ft}}_x \,\|\, p^{\text{stu}}_x\right).
\end{aligned}
\end{equation}
Finally, we combine the two losses using $\rho_t$ from Equation~\ref{eq:rho}:
\begin{equation}
\mathcal{L}_{\text{distill}}
=
\rho_t\,\mathcal{L}^{\text{pre}}_{\text{distill}}
+
(1-\rho_t)\,\mathcal{L}^{\text{ft}}_{\text{distill}}.
\end{equation}

\subsection{History Enhancement}

To capture query-relevant recurring patterns, especially periodic and high-frequency events, we maintain a selective memory for each query pair $(s,r)$. The memory module is implemented as a cache over historical facts $(s,r,o)$, where each entry stores two statistics: its most recent occurrence time and its occurrence frequency. Our update strategy is inspired by adaptive replacement ideas \cite{ARC} and is summarized in Algorithm~\ref{alg:SleMem}.

\begin{algorithm}[t]
\caption{Selective Memory Algorithm}
\label{alg:SleMem}
\KwIn{capacity $C$, historical event sequence for (s,r): $A = [(o_1,t_1),(o_2,t_2),\dots]$}
\KwOut{Cache contents}
Initialize $T_1$ as LRU, $T_2$ as LFU cache, and $B_1, B_2$ as double-ended queues; $p \gets 0$\;
\For{each access $(o_i, t_i)$ in $A$}{
    \tcc{Cache hit}
    \If{$o_i \in T_1$}{
        Move $(o_i,t_i)$ from $T_1$ to $T_2$; set $\mathrm{freq}(o_i)\gets 2$;
    }
    \ElseIf{$o_i \in T_2$}{
        Increase $\mathrm{freq}(o_i)$; update timestamp $t_i$;
    }
    \textbf{Return}
  
    \tcc{Cache miss}
    \If{$o_i \in B_1$}{
      $p \leftarrow \min(C,\,p + \max(|B_2|/|B_1|, 1))$\;
      Call Replace($o_i$)\;
      Remove $o_i$ from $B_1$, insert $(o_i: t_i)$ at the front of $T_2$; 
      set $\mathrm{freq}(o_i)\gets 2$;
    }\ElseIf{$o_i \in B_2$}{
      $p \leftarrow \max(0,\,p - \max(|B_1|/|B_2|, 1))$\;
      Call Replace($o_i$)\;
      Remove $o_i$ from $B_2$, insert $(o_i: t_i)$ at the front of $T_2$; 
      set $\mathrm{freq}(o_i)\gets 2$
    }\Else{
      \If{$|T_1| + |B_1| = C$}{
        \If{$|T_1| < C$}{
          Remove item from $B_1$\;
          Call Replace($o_i$)\;
        }
        \Else{
          Remove item from $T_1$\;
        }
      }
      \ElseIf{$|T_1| + |T_2| + |B_1| + |B_2| \ge C$}{
        \If{$|T_1| + |T_2| + |B_1| + |B_2| = 2C$}{
          Remove item from $B_2$\;
        }
        Call Replace($o_i$)\;
      }
      Insert $(o_i: t_i)$ to $T_1$, set $\mathrm{freq}(o_i)\gets 1$;
    }
}

\SetKwFunction{FReplace}{Replace}
\SetKwProg{Fn}{Function}{:}{}
\Fn{\FReplace($o_i$)}{
  \If{$|T_1| \ge 1$ \textbf{and} (($o_i \in B_2$ \textbf{and} $|T_1| = p$) \textbf{or} ($|T_1| > p$))}{
    Remove item from $T_1$ and insert its key to front of $B_1$\;
  }
  \Else{
    Remove item from $T_2$ and insert its key to front of $B_2$\;
  }
}
\end{algorithm}

Following \cite{DyMemR}, we use a dynamic cache capacity. Let $C_0$ be a base capacity and let $N_t$ be the number of distinct cached facts at time $t$. We set
\begin{equation}
C_t = C_0 + \gamma N_t,
\end{equation}
where $\gamma$ is a tunable coefficient. This allows the memory size to grow adaptively with the stream.

\paragraph{Inference-time re-ranking}
Given a query $(s,r,?,t)$, we retrieve cached candidates $o$ associated with the same query pair $(s,r)$ and compute two priors. Let $\Delta t^o$ denote the time gap since the most recent occurrence of $(s,r,o)$ in memory, and let $\Delta t^{\max}$ denote the maximum such gap among the retrieved candidates. We define the recency prior and frequency prior as
\begin{equation}
\psi^{\text{time}}_{(s,r,t)}(o)=\exp\!\left(1-\frac{\Delta t^o}{\Delta t^{\max}}\right),
\label{eq:his_time}
\end{equation}
and
\begin{equation}
\psi^{\text{freq}}_{(s,r,t)}(o)=\log\!\left(\mathrm{freq}(o)+1\right),
\label{eq:his_freq}
\end{equation}
where $\mathrm{freq}(o)$ is the occurrence count of $(s,r,o)$ stored in memory. For candidates not found in memory, both priors are set to zero. The final history-enhanced score is
\begin{equation}
\hat{\psi}_{(s,r,?,t)}(o)
=
\psi_{(s,r,?,t)}(o)
+
\psi^{\text{time}}_{(s,r,t)}(o)
+
\psi^{\text{freq}}_{(s,r,t)}(o),
\label{eq:his_enh}
\end{equation}
where $\psi_{(s,r,?,t)}(o)$ denotes the student score and $\hat{\psi}_{(s,r,?,t)}(o)$ is used for ranking.

\subsection{Training and Inference}

HiTS-CL consists of an offline pretraining phase followed by continual learning over streaming snapshots. Pretraining follows the standard procedure of the underlying TKGR backbone and produces the pretrained teacher $f^{\text{pre}}_{\theta}$.

At each continual learning stage $t$, we first update the fine-tuned teacher on the current snapshot $\mathcal{G}_t$ and use $\mathcal{G}_{t-1}$ for early stopping. We then train the student on $\mathcal{G}_t$ using both prediction supervision and multi-teacher distillation. The overall objective is
\begin{equation}
\mathcal{L}=\mathcal{L}_{\text{pred}}+\beta\,\mathcal{L}_{\text{distill}},
\end{equation}
where $\beta$ controls the strength of distillation. Both optimization steps use only current or past snapshots, preserving the no-leakage extrapolation protocol.

At inference time, we rank entities using the distilled student score and then apply memory-based re-ranking with the recency and frequency priors in Equation~\ref{eq:his_enh}. In this way, HiTS-CL combines continually updated model parameters for current dynamics and stable knowledge with lightweight historical priors for recurring evidence.

\begin{table*}[!t]
\caption{Statistics of the datasets. All facts are chronologically split into training, validation, and test sets.}
\label{tab:dataset}
\centering
\renewcommand{\arraystretch}{0.95}
\setlength{\tabcolsep}{5pt}
\resizebox{\textwidth}{!}{
\begin{tabular}{lccccccccc}
    \toprule
    Dataset & Entities & Relations & Snapshots & Train & Valid & Test & Test Snapshots & New Entities & Granularity \\
    \midrule
    ICEWS14      & 7,218  & 230 &   365  &  27,391  &   4,514  &    58,825  &   231  & 2,149  & 24~hours \\
    ICEWS18      & 23,033 & 256 &   304  & 140,714  &  23,320  &   304,524  &   203  & 5,082  & 24~hours \\
    ICEWS05-15   & 10,488 & 251 & 4,017  & 138,576  &  22,959  &   299,794  & 2,725  & 3,289  & 24~hours \\
    GDELT        & 7,691  & 240 & 2,751  & 683,646  & 114,076  & 1,480,683  & 1,688  & 1,304  & 15~mins \\
    \bottomrule
    \label{DataSet_stat}
\end{tabular}}
\end{table*}

\begin{table*}[!t]
\caption{Time-aware filtered MRR and Hits@K on ICEWS14/18, ICEWS05-15, and GDELT. We report each backbone under fixed-prefix training and continual learning variants (+FT, +Replay, +Reg, +HiTS-CL).}
    \begin{center}
        \resizebox{\textwidth}{!}{
        \begin{tabular}{lcccccccccccccccc}
        \toprule
         \multirow{2}{*}[-0.5\dimexpr\aboverulesep+\cmidrulewidth+\belowrulesep\relax]{Model}
        & \multicolumn{4}{c}{ICEWS14}  & \multicolumn{4}{c}{ICEWS18} & \multicolumn{4}{c}{ICEWS05-15} & \multicolumn{4}{c}{GDELT}\\
        \cmidrule(lr){2-5} \cmidrule(lr){6-9} \cmidrule(lr){10-13} \cmidrule(lr){14-17}
         & MRR & H@1 & H@3 & H@10 & MRR & H@1 & H@3 & H@10 & MRR & H@1 & H@3 & H@10 & MRR & H@1 & H@3 & H@10 \\
        
        \midrule
        RE-GCN (2021) & 33.81 & 25.66 & 37.48 & 49.47 & 28.41 & 19.33 & 32.00 & 46.23 & 44.29 & 34.76 & 49.07 & 62.65  & 18.41 & 11.56 & 19.45 & 31.76 \\
        RE-GCN+FT & 40.00 & 30.91 & 44.61 & 57.13 & 31.16 & 21.48 & 35.22 & 50.12 & 49.75 & 39.51 & 55.35 & 69.18 & 21.79 & 13.94 & 23.56 & 37.13 \\
        RE-GCN+Replay & 41.29 & 32.16 & 45.86 & 58.61 & 32.48 & 22.53 & 36.75 & 51.93 & 51.64 & 41.21 & 57.42 & 71.31 & 21.75 & 13.86 & 23.46 & 37.21 \\
        RE-GCN+Reg & 40.10 & 31.10 & 44.55 & 57.02 & 31.67 & 21.75 & 35.77 & 51.23 & 52.38 & 41.99 & 58.09 & \textbf{72.12} & 21.10 & 13.42 & 22.72 & 36.09 \\
        RE-GCN+HiTS-CL & \textbf{43.61} & \textbf{34.61} & \textbf{48.48} & \textbf{60.13} & \textbf{34.34} & \textbf{24.44} & \textbf{38.78} & \textbf{53.52} & \textbf{52.53} & \textbf{42.45} & \textbf{58.18} & 71.56 & \textbf{25.22} & \textbf{16.51} & \textbf{27.83} & \textbf{42.28} \\
        \midrule
        CEN (2022) & 30.45 & 22.81 & 34.31 & 44.58 & 30.41 & 20.99 & 34.37 & 48.85 & 46.34 & 36.58 & 51.50 & 64.88 & 19.32 & 12.12 & 20.52 & 33.42 \\
        CEN+FT & 40.98 & 31.84 & 45.80 & 57.99 & 32.37 & 22.56 & 36.67 & 51.63 & 49.41 & 39.09 & 55.04 & 69.02 & 22.25 & 14.27 & 24.09 & 37.88 \\
        CEN+Replay  & 42.51 & 32.97 & 47.76 & 60.15 & 33.56 & 23.38 & 38.10 & 53.47 & 51.80 & 41.28 & 57.76 & 71.68 & 22.09 & 14.08 & 23.89 & 37.82 \\
        CEN+Reg  & 39.49 & 30.88 & 43.86 & 55.59 & 32.30 & 22.44 & 36.56 & 51.63 & \textbf{52.55} & \textbf{41.97} & \textbf{58.47} & \textbf{72.61} & 20.92 & 13.25 & 22.47 & 35.97 \\
        CEN+HiTS-CL & \textbf{43.49} & \textbf{34.45} & \textbf{48.52} & \textbf{60.02} & \textbf{34.23} & \textbf{24.10} & \textbf{38.93} & \textbf{53.87} & 51.63 & 41.44 & 57.31 & 70.85 & \textbf{23.57} & \textbf{15.20} & \textbf{25.73} & \textbf{40.03} \\
        \midrule
        RETIA (2023) & 37.69 & 28.89 & 42.09 & 54.13 & 30.29 & 20.89 & 34.19 & 48.67 & 46.91 & 37.06 & 52.10 & 65.63 & 19.13 & 12.02 & 20.26 & 33.06 \\
        RETIA+FT & 39.93 & 30.85 & 44.57 & 57.04 & 31.65 & 21.83 & 35.79 & 50.92 & 49.04 & 38.75 & 54.54 & 68.64 & 21.40 & 13.54 & 23.11 & 36.79 \\
        RETIA+Replay & 42.19 & 32.79 & 47.04 & 59.83 & 33.56 & 23.38 & 38.10 & 53.47 & 51.65 & 41.09 & \textbf{57.53} & \textbf{71.62} & 21.39 & 13.50 & 23.06 & 36.92 \\
        RETIA+Reg & 41.93 & 32.58 & 46.67 & 59.54 & 31.72 & 21.98 & 35.76 & 50.81 & 48.84 & 38.56 & 54.30 & 68.42 & 21.34 & 13.51 & 23.05 & 36.70 \\
        RETIA+HiTS-CL & \textbf{44.25} & \textbf{35.18} & \textbf{49.13} & \textbf{61.10} & \textbf{34.16} & \textbf{24.31} & \textbf{38.11} & \textbf{53.99} & \textbf{52.12} & \textbf{42.06} & 57.46 & 71.47 & \textbf{25.48} & \textbf{16.90} & \textbf{27.96} & \textbf{42.26} \\
        \midrule
        LogCL (2024) & 39.25 & 29.14 & 44.06 & 59.27 & 30.81 & 20.52 & 34.73 & 51.45 & 51.64 & 41.13 & 57.57 & 71.90 & 22.75 & 13.85 & 24.42 & 40.77 \\
        LogCL+FT & 43.71 & 33.00 & 49.15 & 64.77 & 33.20 & 22.14 & 37.86 & 55.25 & 56.93 & 45.87 & 63.74 & 77.77 & 23.19 & 14.01 & 25.02 & 41.92 \\
        LogCL+Replay & 46.56 & 35.95 & 52.19 & 67.13 & 34.70 & 23.54 & 39.58 & 56.89 & 59.92 & 49.14 & 66.66 & 79.97 & 24.37 & 14.87 & 26.51 & 43.69 \\
        LogCL+Reg & 46.64 & 35.69 & 52.40 & 67.97 & 33.56 & 22.14 & 38.24 & 56.75 & 59.72 & 48.82 & 66.54 & 80.01 & 21.44 & 12.41 & 23.07 & 39.84 \\
        LogCL+HiTS-CL & \textbf{49.01} & \textbf{38.88} & \textbf{54.44} & \textbf{68.55} & \textbf{37.56} & \textbf{26.64} & \textbf{42.54} & \textbf{58.99} & \textbf{60.94} & \textbf{50.40} & \textbf{67.48} & \textbf{80.79} & \textbf{28.54} & \textbf{18.53} & \textbf{31.63} & \textbf{48.50} \\
        \midrule
        DiMNet (2025) & 33.76 & 25.07 & 37.66 & 50.09 & 27.21 & 17.86 & 30.68 & 45.85 & 43.78 & 33.78 & 48.97 & 62.62 & 19.25 & 12.00 & 20.41 & 33.40 \\
        DiMNet+FT & 33.55 & 24.62 & 37.57 & 50.68 & 26.66 & 17.52 & 29.96 & 45.03 & 46.43 & 35.75 & 51.94 & 66.90 & 19.76 & 12.36 & 21.07 & 34.28 \\
        DiMNet+Replay & 37.12 & 27.98 & 41.41 & 54.62 & 28.74 & 19.18 & 32.44 & 47.79 & 48.22 & 37.44 & 53.95 & 68.76 & 20.59 & 12.86 & 22.10 & 35.46 \\
        DiMNet+Reg & 36.42 & 27.33 & 40.46 & 53.88 & 29.05 & 19.38 & 32.72 & 48.38 & 47.35 & 36.77 & 52.91 & 67.52 & 20.95 & 13.18 & 22.54 & 36.01 \\
        DiMNet+HiTS-CL & \textbf{41.50} & \textbf{32.80} & \textbf{46.03} & \textbf{57.47} & \textbf{32.82} & \textbf{23.32} & \textbf{36.98} & \textbf{51.19} & \textbf{53.08} & \textbf{42.95} & \textbf{58.78} & \textbf{72.02} & \textbf{24.61} & \textbf{15.79} & \textbf{27.18} & \textbf{41.89} \\
        \midrule
        \end{tabular}}
    \end{center}
    \label{tab:MainResults1}
\end{table*}

\section{Experiments}

\subsection{Experimental Setup}

\subsubsection{Datasets and Baselines.}

We evaluate \emph{extrapolative} TKGR on four widely used benchmarks: ICEWS14, ICEWS18, ICEWS05-15, and GDELT. We use a chronological 30\%/5\%/65\% split for training, validation, and test, deliberately adopting a longer test horizon to study long-horizon performance under streaming deployment. Detailed dataset statistics are provided in Table~\ref{DataSet_stat}. The ``New Entities'' column indicates the number of entities that appear exclusively in the test set, and the ``Test Snapshots'' column reports the number of temporal snapshots in the test set. In our setting, the earliest period is used for offline pretraining, the subsequent period for validation, and the most recent period is treated as a streaming deployment phase, where the model is continually updated and evaluated snapshot by snapshot in chronological order.

We instantiate HiTS-CL on five representative TKGR backbones: RE-GCN\footnote{\url{https://github.com/Lee-zix/RE-GCN}}~\cite{RE-GCN}, CEN\footnote{\url{https://github.com/Lee-zix/CEN}}~\cite{CEN}, RETIA\footnote{\url{https://github.com/CGCL-codes/RETIA}}~\cite{RETIA}, LogCL\footnote{\url{https://github.com/WeiChen3690/LogCL}}~\cite{LogCL}, and DiMNet\footnote{\url{https://github.com/hhdo/DiMNet}}~\cite{DiMNet}. These models span diverse architectures, including snapshot-based GNNs, temporal sequence encoders, and contrastive learning frameworks. We also compare against DGAR~\cite{TKGC_cl_4}, a continual learning method originally proposed for temporal interpolation and adapted here to our extrapolation protocol. DGAR mitigates forgetting by combining history-context prompting, diffusion-based historical distribution generation, and hierarchical adaptive replay. Specifically, DGAR follows an interpolation setting because its training, validation, and test facts are drawn from the same snapshot, and each task is evaluated only after completing incremental training over all tasks, rather than immediately under a strict temporal order.

\begin{figure*}[!t]
\centering
\begin{adjustbox}{width=0.99\textwidth}
    \begin{minipage}{\textwidth}
\subfloat{\includegraphics[width=0.19\textwidth]{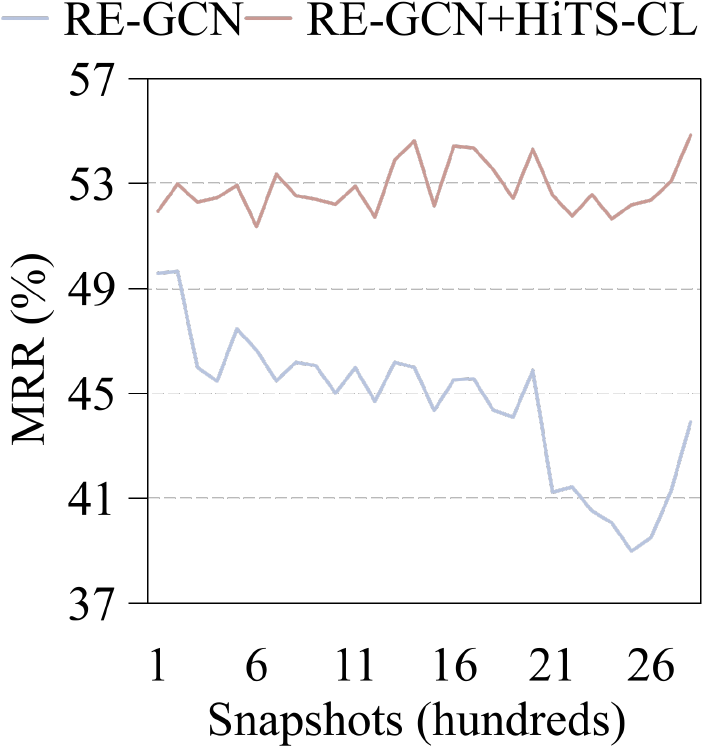}}
\hfill
\subfloat{\includegraphics[width=0.19\textwidth]{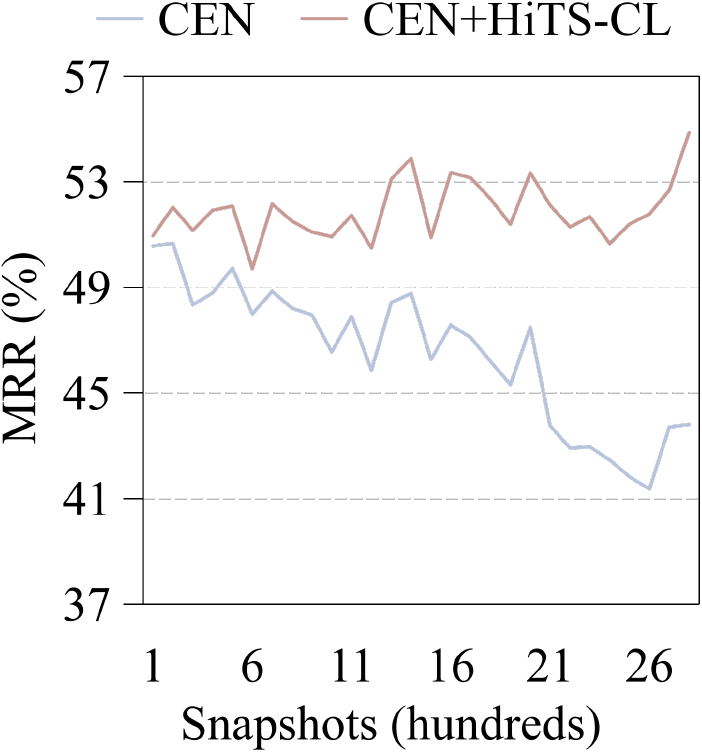}}
\hfill
\subfloat{\includegraphics[width=0.19\textwidth]{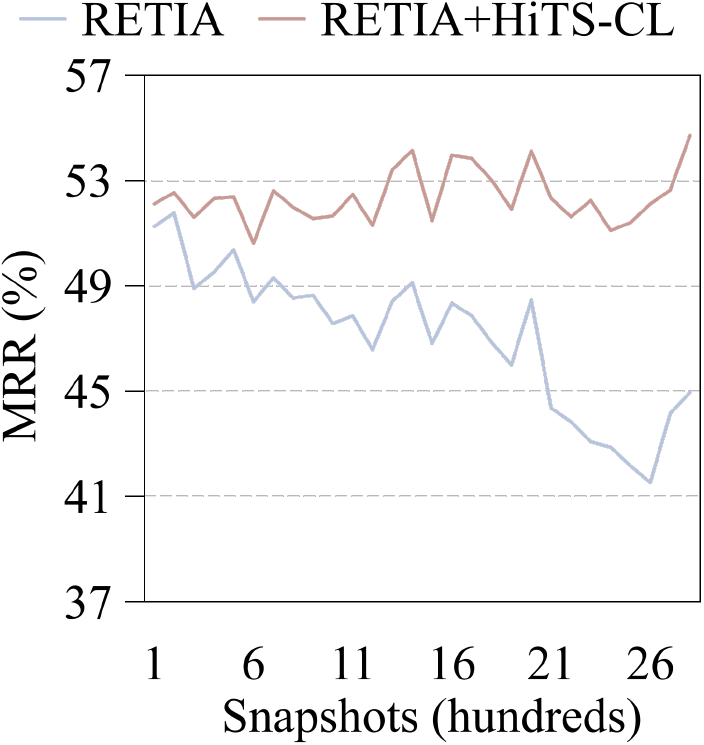}}
\hfill
\subfloat{\includegraphics[width=0.19\textwidth]{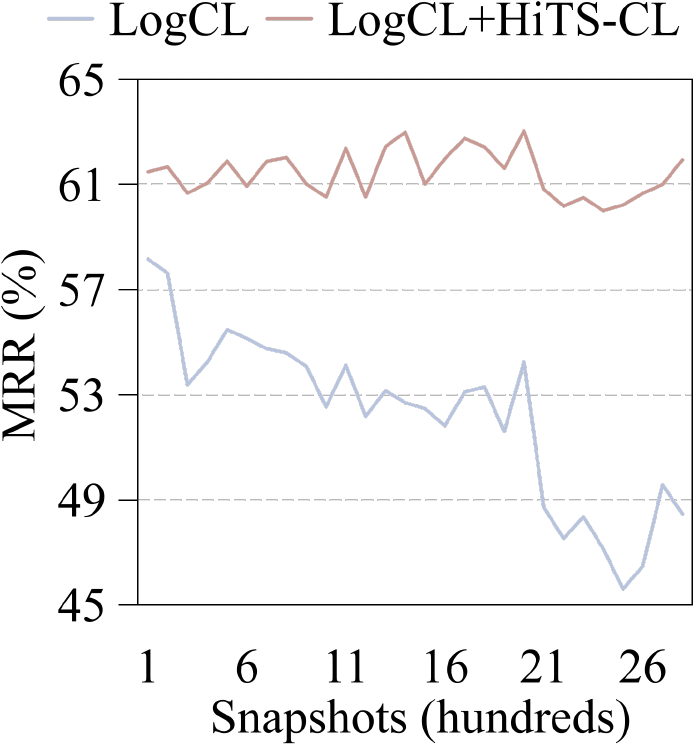}}
\hfill
\subfloat{\includegraphics[width=0.19\textwidth]{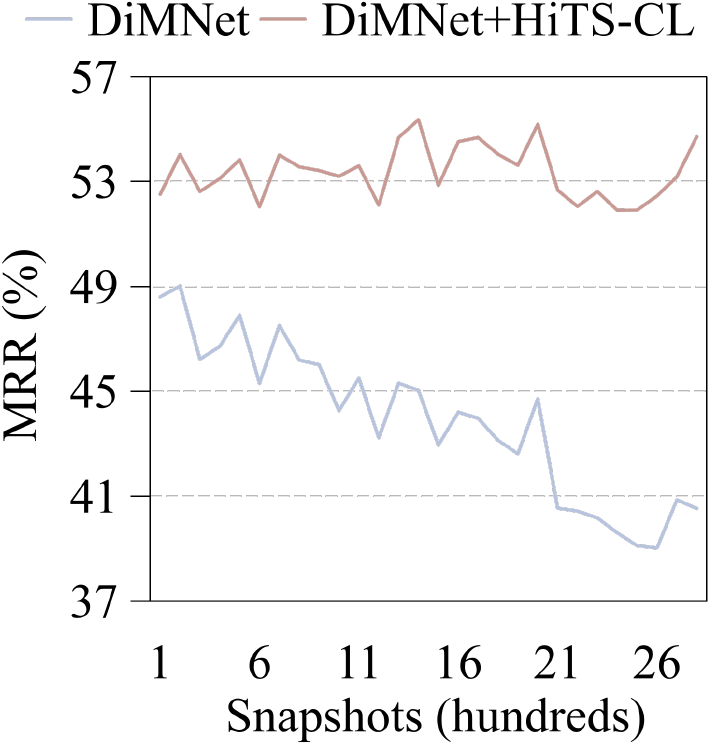}}
\par\medskip
    \end{minipage}
\end{adjustbox}
\caption{MRR degradation trends over time of ICEWS05-15.}
\label{fig:decay_icews0515}
\end{figure*}

\subsubsection{Evaluation Metrics.}
We report standard link prediction metrics, including Mean Reciprocal Rank (MRR) and Hits@K ($K\in\{1,3,10\}$), under the time-aware filtered setting. For each query $(s,r,?,t)$, we rank all candidate entities and filter out other ground-truth entities that form valid facts at the same timestamp.

During evaluation at time $t$, the model is allowed to access only facts with timestamps earlier than $t$ as historical context. This does not leak future information, since facts at the target timestamp are never used as input for queries at $t$. Accordingly, as evaluation proceeds chronologically, earlier validation and test snapshots become available history for later predictions, matching the streaming extrapolation setting.

\subsubsection{Implementation Details}

The training pipeline has two phases: (1) pretraining on the initial training set, and (2) continual learning over the subsequent snapshot stream. During testing, snapshots are processed sequentially by time; after inferring on snapshot $t$, its facts are used as continual learning data for the next step and added to the recurring history memory. Following standard TKGR evaluation, we use a fixed global entity vocabulary constructed from the dataset. The continual learning process is snapshot-based, with early stopping applied based on MRR on the validation snapshot. The early-stopping patience is set to 2. The base capacity $C_0$ and growth factor $\gamma$ of the selective memory are adopted from \cite{DyMemR} with dataset-specific settings. For the +Reg baseline, the regularization coefficient is set to 0.1. All hyperparameters of the base models are kept consistent with their configurations as reported in their official implementation. The same optimizer and learning rate are used throughout both pretraining and continual learning stages. For knowledge distillation, we set the temperature scaling factor $\alpha=2$ and the distillation loss weight $\beta=1$. All experiments are conducted on a Linux server equipped with 52 CPU cores, 384 GB RAM, and an NVIDIA A40 GPU with 48 GB memory. Source code and data are available at \url{https://github.com/liuyansong98/HiTS-CL}

\subsection{Main Results}

\begin{figure}[!t]
    \centering
    \includegraphics[width=\columnwidth]{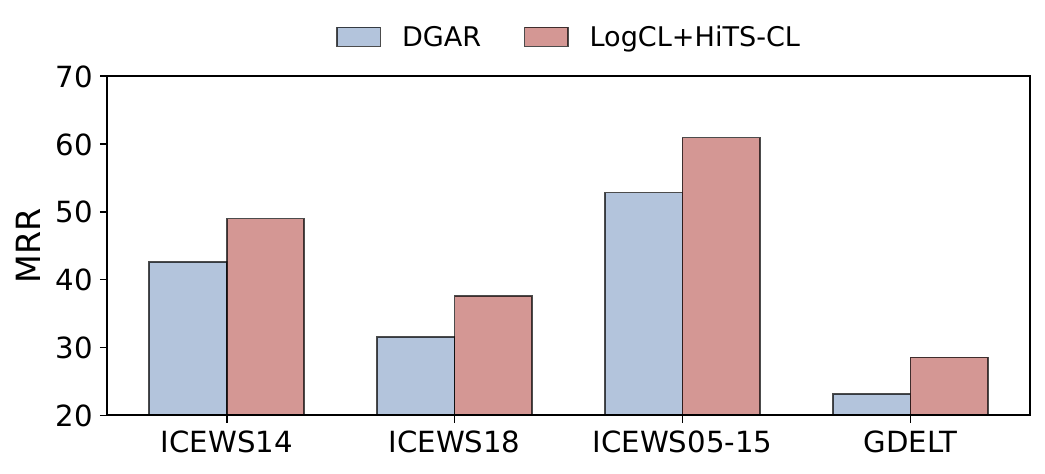}
    \caption{Comparison of DGAR and LogCL+HiTS-CL.}
    \label{fig:dgar_logcl}
\end{figure}

We instantiate HiTS-CL on five representative TKGR backbones and evaluate them on four benchmarks. Table~\ref{tab:MainResults1} compares each backbone under (i) the original static training protocol and (ii) four continual variants: \textbf{+FT}, \textbf{+Replay}, \textbf{+Reg}, and \textbf{+HiTS-CL}. Here, \textbf{+FT} denotes naive continual fine-tuning on the current snapshot, \textbf{+Replay} jointly trains on the current snapshot and a randomly sampled historical snapshot, \textbf{+Reg} penalizes parameter drift between adjacent stages with an $\ell_2$ regularizer, and \textbf{+HiTS-CL} applies our full two-step continual learning framework with history enhancement.

\paragraph{Overall performance}
HiTS-CL consistently achieves the best performance across all datasets and backbones. Compared with the strongest adapted baseline, \textbf{+Replay}, it improves relative MRR by \textbf{6.0\%}, \textbf{6.4\%}, \textbf{2.8\%}, and \textbf{15.7\%} on ICEWS14, ICEWS18, ICEWS05-15, and GDELT, respectively. Relative to the original static models, HiTS-CL yields average MRR gains of \textbf{27.4\%}, \textbf{17.7\%}, \textbf{16.1\%}, and \textbf{29.1\%}, showing that handling non-stationarity is critical for long-horizon extrapolation. 

\begin{figure}[!t]
\centering
\subfloat[\footnotesize ICEWS14]{\includegraphics[width=0.48\columnwidth]{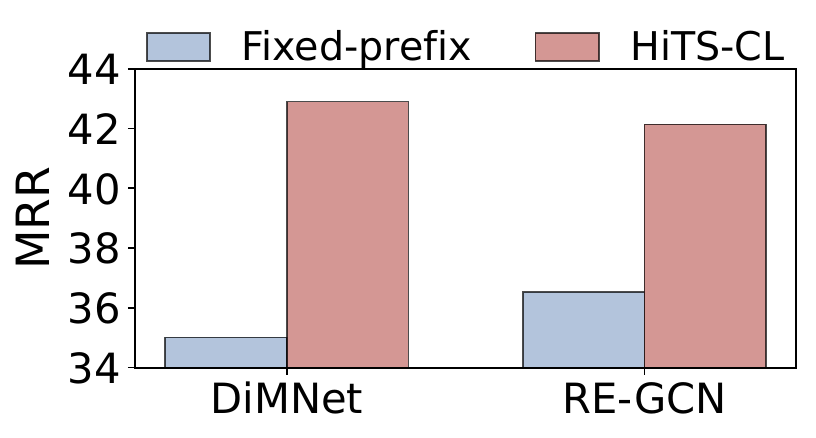}\label{fig:icews14_traditional_vs_hitscl}}
\hfill
\subfloat[\footnotesize ICEWS18]{\includegraphics[width=0.48\columnwidth]{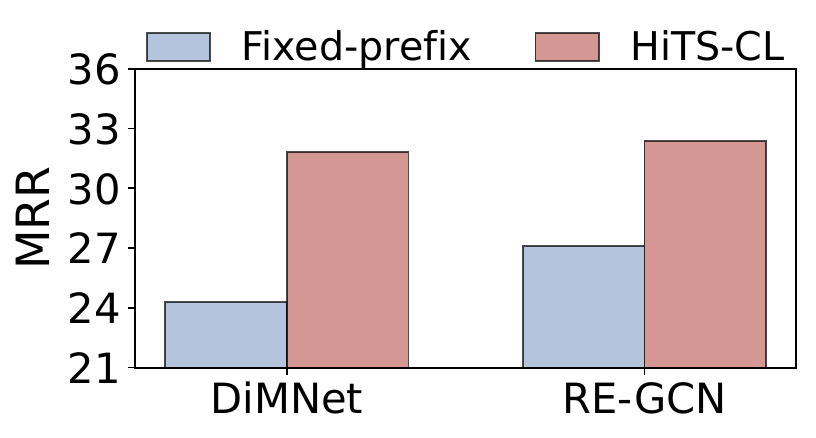}\label{fig:icews18_traditional_vs_hitscl}}
\hfill

\par\bigskip
\subfloat[\footnotesize ICEWS05-15]{\includegraphics[width=0.48\columnwidth]{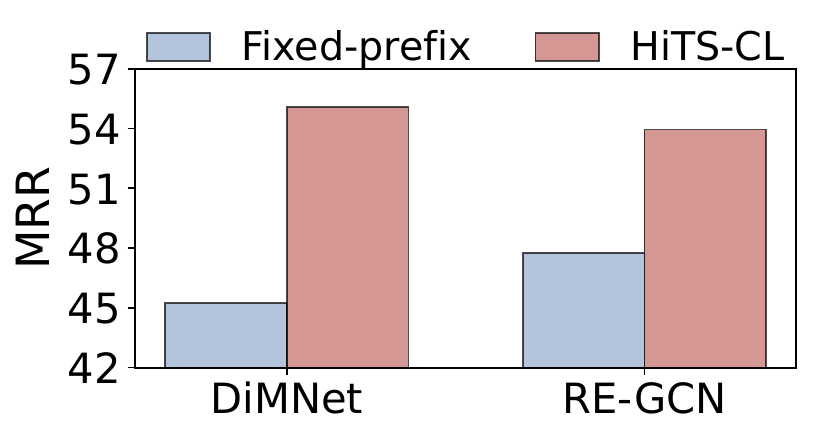}\label{fig:icews05-15_traditional_vs_hitscl}}
\hfill
\subfloat[\footnotesize GDELT]{\includegraphics[width=0.48\columnwidth]{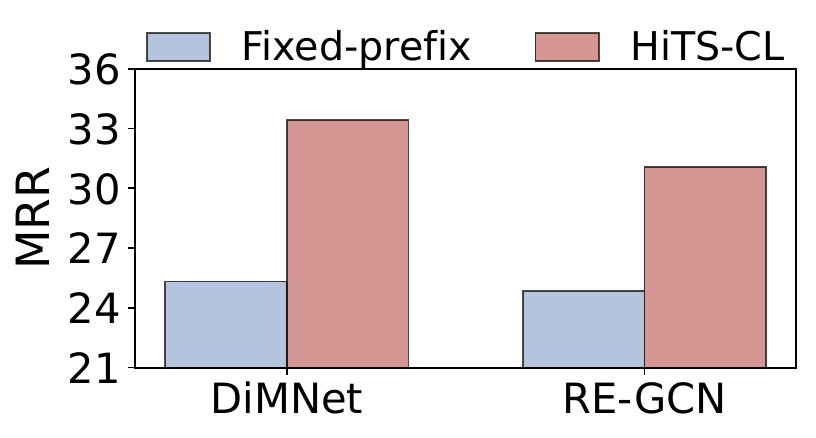}\label{fig:gdelt_traditional_vs_hitscl}}

\caption{Fixed-prefix training versus continual updating under the same evaluation snapshot (MRR).}
    \label{fig:training_comparison}
\end{figure}

\begin{table*}[t!]
\caption{Ablation study on ICEWS14, ICEWS18, ICEWS05-15 and GDELT. All metrics are time-aware filtered.}
    \begin{center}
        \resizebox{\textwidth}{!}{
        \begin{tabular}{lcccccccccccccccc}
        \toprule
         \multirow{2}{*}[-0.5\dimexpr\aboverulesep+\cmidrulewidth+\belowrulesep\relax]{Model}
        & \multicolumn{4}{c}{ICEWS14}  & \multicolumn{4}{c}{ICEWS18} & \multicolumn{4}{c}{ICEWS05-15} & \multicolumn{4}{c}{GDELT}\\
        \cmidrule(lr){2-5} \cmidrule(lr){6-9} \cmidrule(lr){10-13} \cmidrule(lr){14-17}
         & MRR & H@1 & H@3 & H@10 & MRR & H@1 & H@3 & H@10 & MRR & H@1 & H@3 & H@10 & MRR & H@1 & H@3 & H@10 \\
        \midrule
        RE-GCN w/o St.1 & 42.11 & 33.46 & 46.61 & 58.04 & 33.69 & 24.01 & 38.05 & 52.45 & 51.23 & 41.55 & 56.71 & 69.28 & 23.91 & 15.44 & 26.36 & 40.51 \\
        RE-GCN w/o St.2 & 42.47 & 33.60 & 47.33 & 58.74 & 33.54 & 23.83 & 37.91 & 52.32 & 51.59 & 41.51 & 57.24 & 70.58 & 23.09 & 16.43 & 27.69 & 42.03 \\
        RE-GCN w/o His. & 41.52 & 32.36 & 46.19 & 58.60 & 32.20 & 22.36 & 36.31 & 51.54 & 51.19 & 41.03 & 56.69 & 70.46 & 22.07 & 14.14 & 23.85 & 37.59 \\
        RE-GCN+HiTS-CL & \textbf{43.61} & \textbf{34.61} & \textbf{48.48} & \textbf{60.13} & \textbf{34.34} & \textbf{24.44} & \textbf{38.78} & \textbf{53.52} & \textbf{52.53} & \textbf{42.45} & \textbf{58.18} & \textbf{71.56} & \textbf{25.22} & \textbf{16.51} & \textbf{27.83} & \textbf{42.28} \\

        \midrule
        CEN w/o St.1 & 39.58 & 31.10 & 44.33 & 54.94 & 31.95 & 22.20 & 36.33 & 50.93 & 46.80 & 37.38 & 52.24 & 64.22 & 21.07 & 13.05 & 22.93 & 36.86 \\
        CEN  w/o St.2 & 42.89 & 33.24 & 47.24 & 58.94 & 33.34 & 23.51 & 37.79 & 52.51 & 50.25 & 39.98 & 55.98 & 69.66 & 23.43 & 15.13 & 25.57 & 39.74 \\
        CEN  w/o His. & 42.32 & 33.14 & 47.35 & 59.26 & 33.20 & 23.04 & 37.79 & 53.01 & 50.74 & 40.46 & 56.41 & 70.23 & 22.35 & 14.32 & 24.21 & 38.12 \\
        CEN+HiTS-CL & \textbf{43.49} & \textbf{34.45} & \textbf{48.52} & \textbf{60.02} & \textbf{34.23} & \textbf{24.10} & \textbf{38.93} & \textbf{53.87} & \textbf{51.63} & \textbf{41.44} & \textbf{57.31} & \textbf{70.85} & \textbf{23.57} & \textbf{15.20} & \textbf{25.73} & \textbf{40.03} \\

        \midrule
        RETIA w/o St.1 & 43.85 & 34.77 & 48.89 & 60.58 & 33.46 & 23.96 & 37.33 & 52.50 & 50.24 & 41.02 & 55.67 & 67.37 & 25.25 & 16.71 & 27.90 & 41.69 \\
        RETIA w/o St.2 & 42.99 & 33.94 & 47.99 & 59.63 & 33.45 & 23.81 & 37.31 & 52.79 & 51.13 & 40.98 & 56.74 & 70.30 & 25.45 & 16.79 & \textbf{28.15} & 42.19 \\
        RETIA w/o His. & 41.77 & 32.59 & 46.42 & 58.91 & 32.98 & 22.99 & 37.31 & 52.50 & 50.85 & 40.52 & 56.38 & 70.53 & 21.41 & 13.56 & 23.08 & 36.77 \\
        RETIA+HiTS-CL & \textbf{44.25} & \textbf{35.18} & \textbf{49.13} & \textbf{61.10} & \textbf{34.16} & \textbf{24.31} & \textbf{38.11} & \textbf{53.99} & \textbf{52.12} & \textbf{42.06} & \textbf{57.46} & \textbf{71.47} & \textbf{25.48} & \textbf{16.90} & 27.96 & \textbf{42.26} \\
   
        \midrule
        LogCL w/o St.1 & 44.63 & 35.12 & 49.50 & 62.73 & 33.80 & 23.95 & 37.91 & 53.23 & 51.89 & 41.62 & 57.62 & 71.70 & 26.54 & 17.58 & 29.09 & 43.99 \\
        LogCL w/o St.2 & 47.25 & 37.29 & 52.41 & 66.62 & 36.34 & 25.60 & 41.19 & 57.44 & 59.33 & 48.75 & 65.92 & 79.28 & 28.05 & 18.37 & 31.35 & 48.14 \\
        LogCL w/o His. & 46.17 & 35.54 & 51.64 & 66.98 & 34.91 & 23.75 & 39.70 & 57.08 & 59.61 & 48.71 & 66.43 & 79.98 & 24.26 & 14.78 & 26.31 & 48.50 \\
        LogCL+HiTS-CL & \textbf{49.01} & \textbf{38.88} & \textbf{54.44} & \textbf{68.55} & \textbf{37.56} & \textbf{26.64} & \textbf{42.54} & \textbf{58.99} & \textbf{60.94} & \textbf{50.40} & \textbf{67.48} & \textbf{80.79} & \textbf{28.54} & \textbf{18.53} & \textbf{31.63} & \textbf{48.50} \\
        
        \midrule
        DiMNet w/o St.1 & 41.32 & 32.65 & 45.83 & 57.45 & 32.70 & 23.23 & 36.88 & 51.03 & 50.44 & 40.98 & 55.78 & 68.03 & 24.29 & 15.63 & 26.82 & 41.22 \\
        DiMNet w/o St.2 & 39.04 & 30.61 & 43.38 & 54.53 & 31.23 & 22.13 & 35.16 & 48.81 & 51.01 & 40.95 & 56.60 & 69.88 & 24.20 & 15.60 & 26.69 & 41.09 \\
        DiMNet w/o His. & 37.23 & 28.06 & 41.50 & 54.72 & 29.05 & 19.44 & 32.79 & 48.17 & 49.75 & 39.21 & 55.39 & 69.76 & 20.46 & 12.79 & 21.89 & 35.51 \\
        DiMNet+HiTS-CL & \textbf{41.50} & \textbf{32.80} & \textbf{46.03} & \textbf{57.47} & \textbf{32.82} & \textbf{23.32} & \textbf{36.98} & \textbf{51.19} & \textbf{53.08} & \textbf{42.95} & \textbf{58.78} & \textbf{72.02} & \textbf{24.61} & \textbf{15.79} & \textbf{27.18} & \textbf{41.89} \\
        \midrule
        \end{tabular}}
    \end{center}
    \label{tab:Ablation_result}
\end{table*}

\paragraph{Why continual adaptation matters}
Even \textbf{+FT} outperforms static training in nearly all settings, despite suffering from forgetting. This shows that continual adaptation to evolving temporal patterns is already beneficial. However, \textbf{+FT} remains clearly behind \textbf{+Replay} and HiTS-CL, indicating that adaptation alone is insufficient. \textbf{+Reg} is occasionally helpful but lacks consistent gains, suggesting that simple regularization is not enough to balance stability and plasticity under temporal drift.

\paragraph{Comparison with DGAR}
Figure~\ref{fig:dgar_logcl} compares HiTS-CL with DGAR \cite{TKGC_cl_4}, a continual method designed for interpolation and adapted here to extrapolation. Although DGAR improves over static TKGR baselines, it remains consistently weaker than standard TKGR backbones equipped with HiTS-CL. For example, LogCL+HiTS-CL surpasses DGAR by \textbf{11.5\%}, \textbf{11.4\%}, \textbf{11.9\%}, and \textbf{18.4\%} MRR on ICEWS14, ICEWS18, ICEWS05-15, and GDELT, respectively.

\paragraph{Long-horizon robustness}
Figure~\ref{fig:decay_icews0515} plots MRR over time on ICEWS05-15, with each point averaged over 100 snapshots. Static models show clear performance decay as the horizon grows, whereas HiTS-CL substantially stabilizes this trend and often improves performance over time, indicating stronger robustness to temporal drift. 

\noindent\textbf{Takeaway.}
HiTS-CL consistently improves extrapolation accuracy and substantially mitigates long-horizon degradation. Generic continual baselines alleviate part of the decay, but still fall short of HiTS-CL, highlighting the value of jointly modeling current dynamics, stable knowledge, and recurring historical evidence.

\subsection{Training Paradigm Comparison}

We compare continual learning with the traditional fixed-prefix regime by evaluating both on the first snapshot immediately after the training period.

\paragraph{Fixed-prefix training}

For GDELT and ICEWS05-15, we train on snapshots 1--1000, validate on 1002--1101, and test on 1001. For ICEWS14 and ICEWS18, we analogously use 1--100/102--111/101 for train/validation/test.

\paragraph{Continual learning}
For GDELT and ICEWS05-15, we pretrain on 1--900, validate on 901--1000, continually update on 901--1000, and test on 1001. For ICEWS14 and ICEWS18, we analogously use 1--90/91--100/91--100/101 for pretraining/validation/update/test.

As shown in Figure~\ref{fig:training_comparison}, continual learning yields a stronger backbone than fixed-prefix training on the same immediately following test snapshot.

\subsection{Ablation Study}

\begin{table}[!t]
\centering
\caption{Effect of validation strategy on two backbone models (MRR). Using $G_{t-1}$ for validation consistently performs slightly better than holding out 15\% $G_t$.}
\resizebox{\columnwidth}{!}{
\begin{tabular}{llcccc}
\toprule
Backbone & Validation & ICEWS14 & ICEWS05-15 & GDELT & ICEWS18 \\
\midrule
\multirow{2}{*}{DiMNet}
 & $G_{t-1}$ & \textbf{41.50} & \textbf{53.08} & \textbf{24.61} & \textbf{32.82} \\
 & $G_t$ (15\%) & 41.03 & 52.49 & 24.36 & 32.69 \\
\midrule
\multirow{2}{*}{RE-GCN}
 & $G_{t-1}$ & \textbf{43.61} & \textbf{52.53} & \textbf{25.22} & \textbf{34.34} \\
 & $G_t$ (15\%) & 43.29 & 51.48 & 25.14 & 33.43 \\
\bottomrule
\end{tabular}}
\label{tab:Validation_Strategy_Analysis}
\end{table}

To quantify the contribution of each component, we conduct ablations on four datasets and multiple TKGR backbones. We evaluate three modules: continual fine-tuning (\textbf{St.1}), multi-teacher distillation (\textbf{St.2}), and history enhancement (\textbf{His}). Table~\ref{tab:Ablation_result} reports evaluation results across all four datasets.

\paragraph{Effect of continual fine-tuning (St.1)}
Removing St.1 causes the largest performance drop across settings. Without continual fine-tuning, the model cannot track current dynamics, and the framework relies primarily on the pretrained teacher, weakening adaptation to the current regime.

\paragraph{Effect of multi-teacher distillation (St.2)}
Removing St.2 also consistently degrades performance. This shows that adapting to recent snapshots alone is insufficient: distillation is needed to combine stable knowledge from the pretrained teacher with current dynamics captured by the fine-tuned teacher, thereby reducing forgetting.

\paragraph{Effect of history enhancement (His)}
Disabling His further reduces performance, with the largest drops on GDELT. This confirms that recurring historical evidence provides complementary signals beyond recent snapshots, especially when observations are sparse or irregular. The magnitude of the drop varies across backbones, suggesting different sensitivity to historical cues.

\noindent\textbf{Takeaway.}
All three components are important: St.1 captures current dynamics, St.2 preserves stable knowledge while reducing forgetting, and His provides up-to-date recurring historical evidence.

\subsection{Validation Strategy Analysis}
\label{sec:Validation_Strategy}

\begin{table}[t]
\centering
\caption{Inference-time computational cost of the History Enhancement (His) module. 
We report the average runtime per snapshot and the memory usage of the historical cache on four datasets.}
\resizebox{\columnwidth}{!}{
\begin{tabular}{lcccc}
\toprule
 & ICEWS14 & ICEWS05-15 & ICEWS18 & GDELT \\
\midrule
Runtime Per snapshot (s) & 0.064 & 0.032 & 0.408 & 0.297 \\
Memory usage (MB) & 134.18 & 339.35 & 535.66 & 965.93 \\
\bottomrule
\end{tabular}}
\label{tab:his_cost}
\end{table}

\begin{figure}[!t]
    \centering
    \includegraphics[width=\columnwidth]{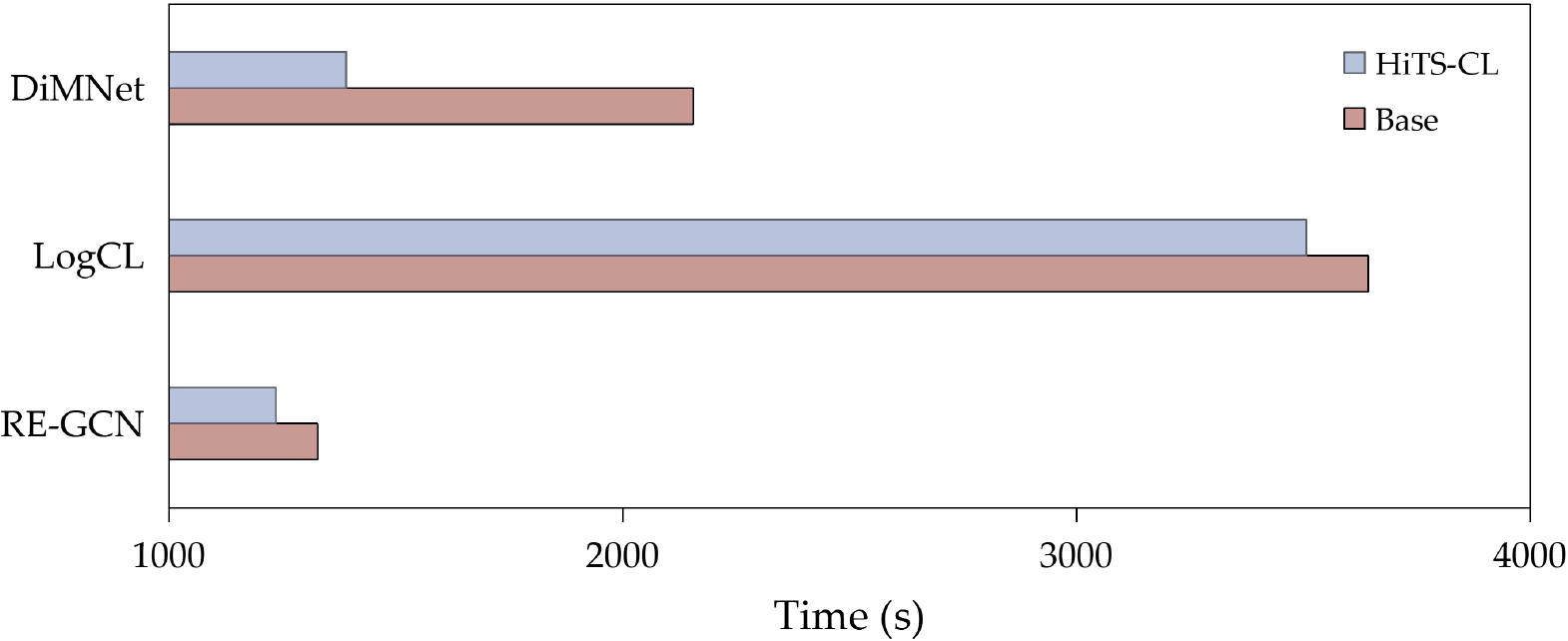}
    \caption{Runtime comparison (in seconds) between full training and continual learning.}
    \label{fig:training_time}
\end{figure}

In our continual extrapolation setting, the model at stage $t$ is updated using the newly arrived snapshot $\mathcal{G}_t$ while avoiding information leakage from future timestamps. Our default strategy trains on the full $\mathcal{G}_t$ and validates on the previous snapshot $\mathcal{G}_{t-1}$. A natural alternative is to hold out part of $\mathcal{G}_t$ as a validation set.

To examine this design choice, we compare two strategies: (1) validating on $\mathcal{G}_{t-1}$ (default), and (2) reserving $15\%$ of $\mathcal{G}_t$ for validation while using the remaining $85\%$ for training. Table~\ref{tab:Validation_Strategy_Analysis} reports the results on two representative backbones (DiMNet and RE-GCN) across four datasets.

Validation on $\mathcal{G}_{t-1}$ consistently yields slightly better performance than validation on $\mathcal{G}_t$. This indicates that holding out part of the current snapshot weakens the training signal for learning newly emerging temporal patterns. In contrast, validating on $\mathcal{G}_{t-1}$ preserves the full training signal and provides a stable selection criterion, as adjacent snapshots are temporally close and typically share similar distributions. Moreover, performance improvements on $\mathcal{G}_{t-1}$ after learning from $\mathcal{G}_t$ are consistent with \emph{backward transfer} in continual learning.

These results support our default validation strategy.

\subsection{Efficiency Analysis}

\subsubsection{Inference Efficiency}

During inference, only the distilled student model is used, while the pretrained and fine-tuned teacher models are employed solely for distillation during continual updates. 
As a result, the inference cost is the same as that of the backbone, with the only additional overhead coming from the History Enhancement module. 
To quantify this overhead, we measure the average runtime per snapshot and the memory usage of the historical cache on four datasets. 
The results are summarized in Table~\ref{tab:his_cost}. 
The History Enhancement module introduces minimal computational cost (e.g., 0.064s per snapshot on ICEWS14) and moderate memory usage, remaining below 1GB even on GDELT. 
These findings indicate that HiTS-CL achieves efficient inference while incorporating history-based reasoning, making it practical for large-scale TKGR streams.

\subsubsection{Training Efficiency}

To assess the training efficiency of HiTS-CL, we compare its training time on ICEWS14 using three representative backbones: RE-GCN, LogCL, and DiMNet. This efficiency comparison is conducted under a separate controlled setting, where both methods are evaluated after observing the same first 80\% of the timeline. Under the fixed-prefix protocol, each model is trained once on the first 80\% of the data. In contrast, HiTS-CL first pretrains on the first 30\% and then performs continual learning on the subsequent 50\% (from 30\% to 80\%).

Figure~\ref{fig:training_time} compares the overall training time of the original fixed-prefix models and their HiTS-CL variants. On average, HiTS-CL reduces total training time by \textbf{15.4\%}. This gain comes from replacing one expensive full training run with stage-wise updates on smaller increments, which typically require fewer epochs and converge faster.

\begin{table}[!t]
\centering
\caption{Training time breakdown of HiTS-CL (seconds) on ICEWS14, including pretraining, continual fine-tuning, and distillation stages.}
\label{tab:training_time}
\resizebox{\columnwidth}{!}{
\begin{tabular}{lcccc}
\toprule
Model & Pretrain & Fine-Tune & Distillation & Overall \\
\midrule
DiMNet & 620 & 279 & 492 & 1,391 \\
LogCL  & 1,618 & 736 & 1,153 & 3,507 \\
RE-GCN & 526 & 183 & 527 & 1,236 \\
\bottomrule
\end{tabular}}
\end{table}

\begin{table}[t]
\centering
\scriptsize
\caption{Case studies illustrating the effect of the History Enhancement module.
Entities in bold denote the ground-truth entity.}
\resizebox{\columnwidth}{!}{
\begin{tabular}{ll}
\toprule
Query &
(Media\_(China), Make\_statement, ?, 299) \\
\midrule
HiTS-CL w/o His. (Top-5) &
Japan, South\_Korea, Kim\_Jong-Un,
\textbf{China}, North\_Korea \\

\midrule
HiTS-CL (Top-5) &
\textbf{China}, Japan,
South\_Korea, Kim\_Jong-Un, North\_Korea \\

\midrule
Historical Entities &
\textbf{China (freq=11, latest=261)}, Japan (freq=1, latest=222) \\
\bottomrule
\end{tabular}}
\label{tab:case_study}
\end{table}

To further clarify the computational cost of HiTS-CL, Table~\ref{tab:training_time} reports a detailed breakdown of its training time into three stages: pretraining, continual fine-tuning, and distillation. We observe that pretraining and distillation account for the majority of the total cost, while the continual fine-tuning stage is relatively lightweight across all three backbones. For example, on LogCL, the runtime is 1,618s for pretraining, 736s for continual fine-tuning, and 1,153s for distillation, giving a total of 3,507s. Similar patterns are observed for RE-GCN and DiMNet.

Overall, these results show that HiTS-CL is not only effective but also computationally practical for long-term TKGR, with manageable overhead in each stage and lower overall training time than fixed-prefix training.

\subsection{Case Study}

We further present a case study to illustrate how the History Enhancement module improves prediction quality. 
Table~\ref{tab:case_study} compares the top-ranked candidate entities predicted by HiTS-CL with and without the History Enhancement module. 
In the example, entities that appear frequently in historical events and have recent occurrences (e.g., \textbf{China}) are promoted to higher ranks after applying the History Enhancement module. 
This indicates that incorporating simple historical statistics, such as frequency and recency, can effectively guide the model toward temporally consistent predictions.

\section{Hyperparameter Sensitivity Analysis}

\begin{figure}[!t]
\centering
\subfloat[\footnotesize ICEWS14]{\includegraphics[width=0.45\columnwidth]{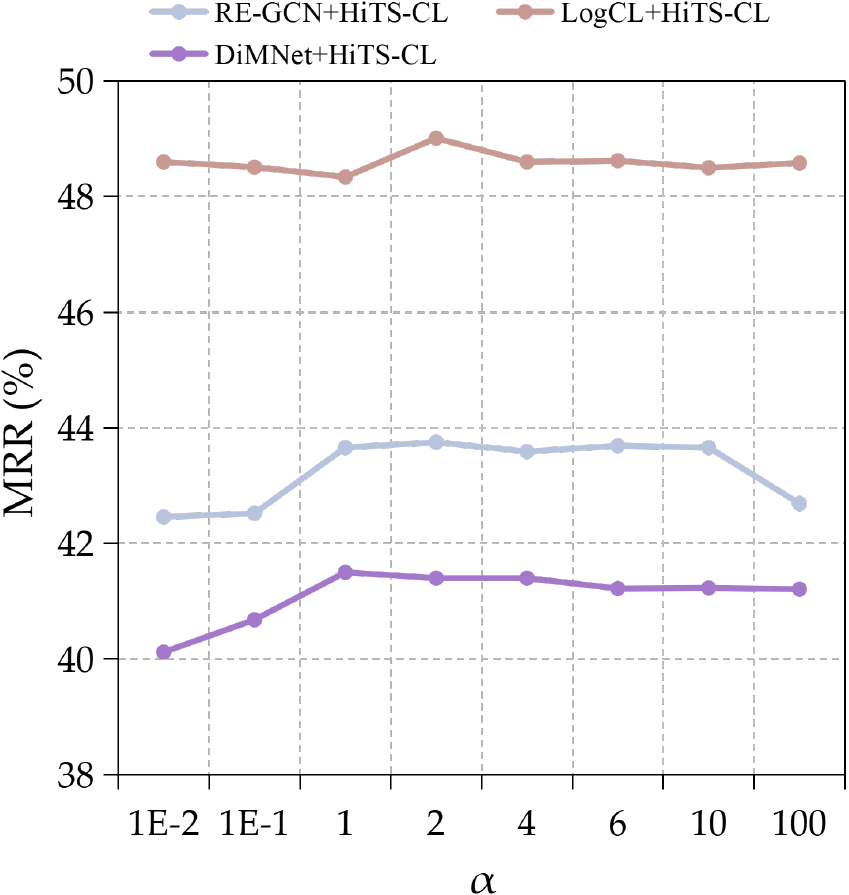}\label{fig:parameter_ICE14_temp}}
\hfill
\subfloat[\footnotesize ICEWS14]{\includegraphics[width=0.45\columnwidth]{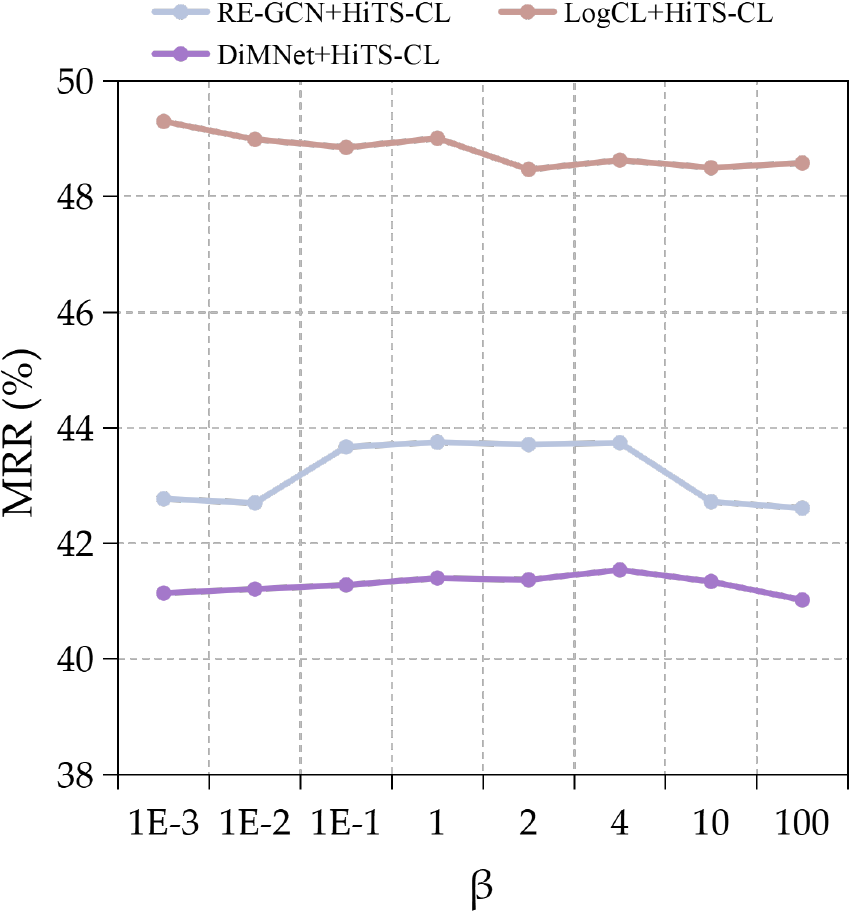}\label{fig:parameter_ICE14_weight}}
\hfill

\par\bigskip
\subfloat[\footnotesize GDELT]{\includegraphics[width=0.45\columnwidth]{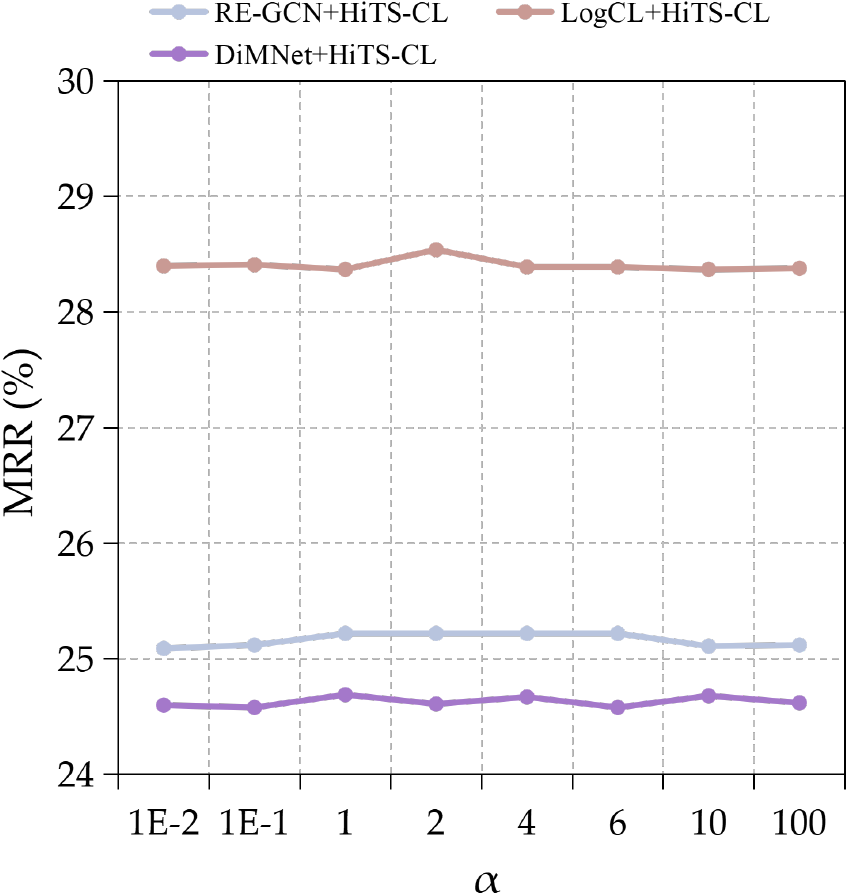}\label{fig:parameter_GDELT_temp}}
\hfill
\subfloat[\footnotesize GDELT]{\includegraphics[width=0.45\columnwidth]{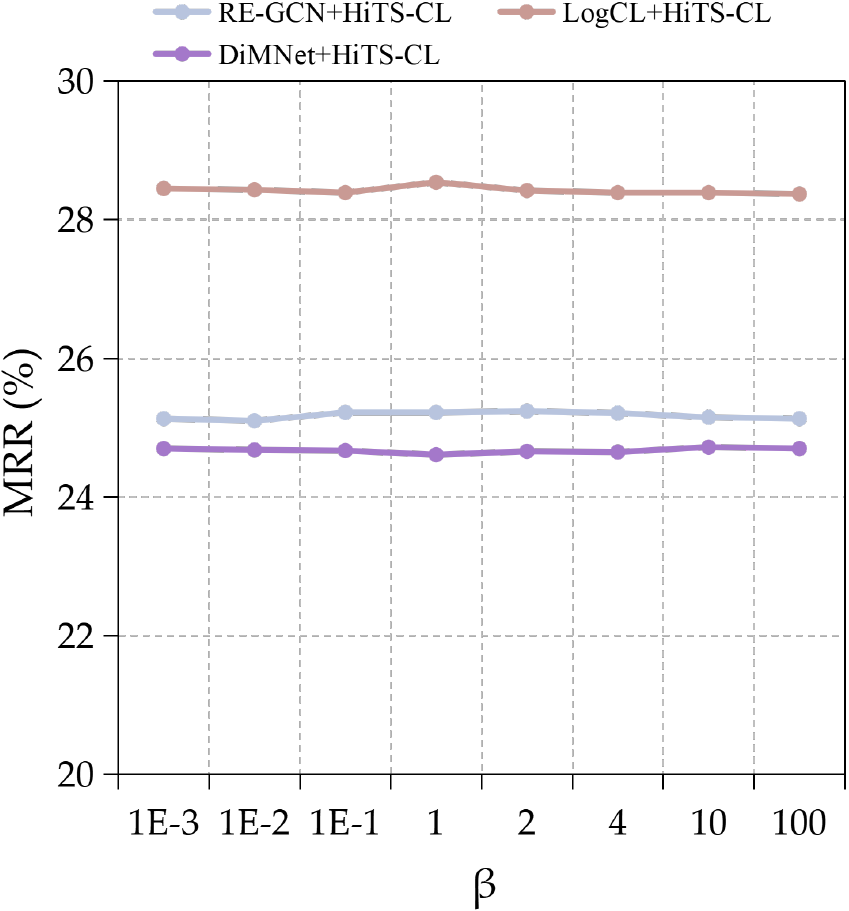}\label{fig:parameter_GDELT_weight}}
\caption{Sensitivity analysis of $\alpha$ and $\beta$ hyper-parameters.}
\label{fig:parameter_sensitivity}
\end{figure}

In this section, we investigate the impact of two hyperparameters—the temperature scaling factor $\alpha$ and the distillation loss weight $\beta$—on model performance using RE-GCN, LogCL, and DiMNet as backbones on the ICEWS14 and GDELT datasets. The results are shown in Figure~\ref{fig:parameter_sensitivity}. 

By adjusting $\alpha$, the model can amplify or attenuate the differentiation among samples with varying confidence levels during distillation. As shown in Figure~\ref{fig:parameter_ICE14_temp} and \ref{fig:parameter_GDELT_temp}, experimental results on both datasets demonstrate that the model is generally robust across a wide range of $\alpha$ values. However, extremely high or low values can lead to performance degradation. The results for the distillation loss weight $\beta$ are shown in Figure~\ref{fig:parameter_ICE14_weight} and \ref{fig:parameter_GDELT_weight}. This parameter exhibits different behavior across the datasets: it shows broad robustness on GDELT, but some sensitivity is observed for certain base models on the ICEWS14 dataset.

\section{Conclusion}
We revisit extrapolative temporal knowledge graph reasoning and argue that the prevailing fixed-prefix training paradigm is fundamentally mismatched to non-stationary temporal streams, which leads to substantial long-horizon performance degradation. To address this issue, we formulate extrapolative TKGR as continual learning over streaming snapshots and propose HiTS-CL, a history-enhanced two-step continual learning framework. HiTS-CL captures current dynamics through continual fine-tuning, preserves stable knowledge through multi-teacher adaptive distillation, and incorporates recurring historical evidence through selective memory. Experiments on five TKGR backbones across four benchmarks show that HiTS-CL consistently improves extrapolation accuracy, mitigates long-horizon degradation, and outperforms strong continual-learning baselines as well as a representative continual TKG method under a unified no-leakage extrapolation protocol. These results support continual learning as a practical and effective paradigm for long-running temporal knowledge graph streams.

\begin{acks}
This work is supported in part by the National Natural Science Foundation of China (NSFC, Grant Nos. 72571019, 62006005, and 62372054), the National Key Research and Development Program of China (Grant No. 2022YFC3302200), the Ministry of Education Humanities and Social Sciences Project (Grant No. 22YJCZH242), and the Special Fund of the Ministry of Science and Technology of China for Major Scientific Research Facilities and Large-scale Research Instruments (Project: National Network Management Platform System Operation and Support Service).
\end{acks}

\section*{GenAI Usage Disclosure}
The authors used generative AI tools only for language polishing and grammar checking. All technical ideas, experimental design, implementation, analysis, and conclusions were developed and verified by the authors.

\bibliographystyle{ACM-Reference-Format}
\balance
\bibliography{sample-base}

\end{document}